# AI-ASSISTED GERMAN EMPLOYMENT CONTRACT REVIEW: A BENCHMARK DATASET

Oliver Wardas / Florian Matthes

Research Associate, Technical University of Munich, School of CIT, Software Engineering for Business Information Systems
Boltzmannstrasse 3, 85748 Garching, DE
oliver.wardas@tum.de; https://wwwmatthes.in.tum.de

Professor, Technical University of Munich, School of CIT, Software Engineering for Business Information Systems
Boltzmannstrasse 3, 85748 Garching, DE
matthes@tum.de; https://wwwmatthes.in.tum.de

*Abstract:* *Employment contracts are used to agree upon the working conditions between employers and employees all over the world. Understanding and reviewing contracts for void or unfair clauses requires extensive knowledge of the legal system and terminology. Recent advances in Natural Language Processing (NLP) hold promise for assisting in these reviews. However, applying NLP techniques on legal text is particularly difficult due to the scarcity of expert-annotated datasets. To address this issue and as a starting point for our effort in assisting lawyers with contract reviews using NLP, we release an anonymized and annotated benchmark dataset for legality and fairness review of German employment contract clauses, alongside with baseline model evaluations.*

## 1. Introduction

Despite an increasing academic interest in Legal NLP research over the last years, AI-assisted contract review, especially in languages other than English, has received little attention [KATZ 2023]. One major hurdle for that may be the scarcity of sufficient, annotated training data. Semantic annotations of legal texts can only be done by legal experts, resulting in high costs and a scarcity of publicly available datasets. The situation worsens when legal texts, such as employment contracts, include sensitive personal information. A partnership with a German law firm specializing in Economic Law now enables us to conduct more research in this area. As part of a collaborative project, we aim to design, implement, and evaluate a prototypical AI-based system for assisting in the review and correction of German employment contracts. To initiate our research efforts and encourage further investigations and experiments by other researchers, we release an anonymized and annotated dataset of clauses from German employment contracts (License: CC BY-NC 4.0), along with their respective legality and categorization labels. Additionally, we provide benchmarks for both open- and closed-source baseline models.

## 2. Background

### 2.1. Employment Contracts

Employment contracts are legally binding agreements between employers and employees that define the terms and conditions of employment. Typically written by the employer, these contracts specify job responsibilities, compensation, work hours, benefits, and conditions for termination or changes. They serve to protect the rights and obligations of both parties, ensuring clarity and preventing disputes. However, changing legislation, repeated use of the same contract templates over extended periods, and modifications to meet individual employee needs can result in legally void clauses — clauses that are unenforceable by law and have no legal effect. If unnoticed, these void clauses can constitute unfavorable conditions for the employee and put the employer at risk of lawsuits if discovered later. Insufficient legal knowledge hinders employers and employees from identifying and replacing void clauses. Consequently, if any employment contract reviews are conducted, they are often times outsourced to external law firms, making the process costly and time-consuming.

## 2.2. Legal NLP

Legal work revolves around writing, reading and interpreting large volumes of text. Modern legal systems continuously produce massive amounts of textual data [KATZ 2020, COUPETTE 2021]. The quantity and complexity [RUHL 2017, RUHL 2015] of legal text makes it a promising field for applying NLP technologies in order to increase efficiency, decrease costs and improve access to justice [CABRAL 2012].

## 3. Related Work

## 3.1. Contract Analysis

Prior to this work, research has been conducted in the field of technology-assisted review of different types of contracts. The studied approaches include the use of AI and Machine Learning as well as rule-based approaches. The following contributions are most related to our work:

[LEE 2019] developed a system aimed at the automated detection of potentially hazardous clauses in construction contracts. This system is specifically designed to function within the confines of contracts that adhere to the FIDIC standard, employing a set of manually crafted rules for the analysis of clauses.

[HENDRYCKS 2021] introduced a comprehensive benchmark dataset containing approximately 13,000 annotated clauses across 25 different contract types. This dataset not only provides a wide range of annotations, including party names and relevant dates, but also identifies clauses with potential risks to operability or revenue.

[LEIVADITI 2020] focused on identifying 17 specific types of risky clauses in leasing contracts and compiled a dataset of 179 English-language documents, complementing it with an ML model, trained on these contracts.

Research efforts within the CLAUDETTE project [LIPPI 2019] and by [BRAUN 2021] included employing machine learning-based methods for assessing English and German general terms and conditions of online shops.

## 3.2. Legal NLP Datasets

Other legal NLP datasets include: [CHALKIDIS 2021, NIKLAUS 2023, BARALE 2022, XIAO 2019, ZHONG 2020, HENDRYCKS 2021, AUMILLER 2022, XIAO 2018, HOLZENBERGER 2020, POUDYAL 2020]

## 4. The Dataset

The dataset[1] contains 1094 samples which consist of a clause, the title of its section and three annotations, including a legality label ('valid', 'unfair' or 'void'), one of 14 categories and, in case of a negative label, a short explanation of what makes the clause void or unfair. A clause, in this case, is a text segment from a German employment contract that describes a semantically self-contained part of the agreement. Its title is the name of the section it is contained in. The legality annotations are created by two lawyers with an inter-annotator agreement of 96.4% regarding the legality labels and 100% regarding the categories. Table 1 shows a sample clause, with its respective title, category, legality classification and explanation.

| |
|---|
| **Clause (de):** Die teilweise oder vollständige Abtretung und Pfändung der Vergütung ist ausgeschlossen.<br>**Translation (en):** The partial or complete assignment and seizure of the remuneration is excluded. |
| **Title (de):** 5 Abtretungen/Pfändungen<br>**Translation (en):** 5 Assignments/Garnishments |
| **Category (de):** Pfändung/Abtretung<br>**Translation (en):** Garnishment/Assignment |
| **Classification:** Void (1) |
| **Explanation (de):** Pfändungs-/Abtretungsverbot seit 10/2021 nach § 308 Nr. 9 lit. a BGB in AGB ausgeschlossen<br>**Translation (en):** Prohibition of garnishment/assignment excluded in general terms and conditions since 10/2021 according to § 308 No. 9 lit. a BGB. |

Table 1: Example for a Void Clause from a German Employment Contract

### 4.1. Contract Source

The contracts were taken from the law firm's client data (terms and conditions cover this type of anonymized data usage). They were first anonymized, then segmented into clauses and finally annotated. Additionally, they were categorized into 14 classes. It is important to mention that the contracts were selected randomly and are, at least in the scope of contracts that clients handed to the law firm, not biased towards more or less problematic (i.e. containing many void or unfair clauses) contracts. Considering that clients are more likely to hand in contracts for review that were not updated for a longer period of time or that they suspect problems with, the dataset might still be biased towards problematic contracts. However, a system trained on this dataset would most likely also be used in a setting where the same bias exists.

### 4.2. Label Explanations

The clauses are classified into one of the following three labels according to their respective criteria. If a clause is unfair and void, it is classified as void, though one can assume that void clauses are most likely also unfair. Clauses classified as unfair cannot directly be subsumed as void under existing regulations and case law. However, the annotators estimate that if those unfair clauses would be challenged in court, there would be a decent chance of them being ruled as void.

---

[1] https://github.com/sebischair/Employment-Contract-Clauses-German

- **Valid** (0): A clause was classified as valid if it is neither unfair nor void. It is the most frequent label (~80%).
- **Unfair** (0.5): A clause was classified as unfair if it introduces major disadvantages for one party while not being void.
- **Void** (1): A clause was classified as void if it fails to comply with regulations or if it can be subsumed as void under existing case law.

### 4.3. Annotation Process

The annotation process included 3 annotation rounds, which were all executed by the same two annotators - lawyers from a small size law firm, based in Germany and specialized on Economic Law. In each annotation round, the clauses were chosen out of multiple randomly selected contracts in a way that they cover all common categories of employment contract clauses. No contracts were reused between different annotation rounds.

In the first round, 100 randomly chosen clauses were each annotated by both lawyers. The inter-annotator agreement on this first iteration was 72.6%. After this initial round, the annotators agreed on a gold-standard for the legality annotations and on 14 categories to classify the clauses in. In a second annotation round they again both annotated 300 clauses. This time the inter-annotator agreement was 96.4% and there was no difference in the category classifications. In a third and final round the lawyers annotated 2 distinct sets of about 400 clauses each, resulting in 794 annotated clauses. The final set contains the results of round two and three, resulting in 1094 samples.

### 4.4. Data Insights

Table 2 shows the data distribution over the selected categories and labels. It shows that the probability for a clause being unfair or void severely differs between the different categories. A main reason for this can be changes in regulation which, if unnoticed by the employer, can leave clauses void. One example for this issue is a change of the regulation § 309 Nr.13 BGB in October 2016. Since this date, employers are not allowed to demand the paper written and hand-signed form for employee requests which have to adhere to cut-off periods. Contracts which do not comply with this changed regulation are the main reason for the high percentage of void or unfair clauses in the categories 'Ausschlussfristen' and 'Form'. Due to copying and re-using parts of employment contracts to create new ones being a common practice, the dataset can contain duplicates in terms of clauses. Since the clauses were annotated by different lawyers, it can also happen that exactly the same clause has different annotations. These cases constitute about 0.2% of the dataset.

| Category (de) | Translation (en) | Valid | Unfair | Void | Sum | Void or Unfair |
|---|---|---|---|---|---|---|
| **Sonstiges** | Other | 161 | 11 | 6 | 178 | 9.6% |
| **Vergütung** | Compensation | 100 | 27 | 21 | 148 | 32.4% |
| **Vertragsstrafe** | Penalty Clause | 111 | 24 | 10 | 145 | 23.5% |
| **Durchführung** | Execution | 111 | 12 | | 123 | 9.8% |
| **Kündigung/Beendigung** | Termination | 93 | 12 | 5 | 110 | 15.5% |
| **Leistungen** | Tasks | 98 | 5 | 1 | 104 | 5.8% |
| **Krankheit** | Illness | 73 | 17 | | 90 | 18.9% |
| **Urlaub** | Vacation | 62 | 12 | | 74 | 16.2% |
| **Verjährung** | Statute of Limitations | 23 | 7 | 6 | 36 | 36.1% |
| **Form** | Format | 21 | 12 | 1 | 34 | 38.2% |
| **Pfändung/Abtretung** | Garnishment | 12 | 9 | 7 | 28 | 67.9% |
| **Überstunden** | Overtime | 9 | 1 | 1 | 11 | 18.2% |
| **Ausschlussfristen** | Cut-off Periods | | 2 | 7 | 9 | 100% |
| **Erfindungen** | Inventions | 4 | | | 4 | 0% |
| **SUM** | | 875 | 149 | 70 | 1094 | 20% |

**Table 2: Data Distribution on 14 Distinct Categories and 3 Legality/Fairness Labels: 'Valid', 'Unfair', 'Void'**

## 5. Baseline Models

In addition to the annotated dataset, we release benchmarking results for some models, serving as the baseline for further research and prediction improvements. We experiment with both prompt engineering and fine-tuning. All used models were already pretrained on large text corpora. We achieve the best performance with fine-tuning closed source models, provided by OpenAI (through the OpenAI API [OPENAI 2024]), though we also experiment with some open source models. We provide precision, recall and f1-score measurements. In all experiments, the input had to be binary classified as either okay or problematic, meaning void or unfair. When possible, additional model layers were placed on top of the pretrained model to produce binary predictions. For the closed source models we used, this was not possible, so we finetuned the models to produce single word predictions *'okay'* or *'problematisch'* (English: 'okay' or 'problematic') which we transformed to binary values afterwards. For our prompt engineering attempts we instructed the model to think step by step (inspired by [WEI 2022]) and conclude its thoughts with a final decision towards either okay or problematic. The full response often times contained these words in its argumentation already, so we derived the binary classification based on which of these words occurred last in the response.

### 5.1. Models and Training Parameters

We only used models that were already pretrained on large text corpora and were either available through an API or which could be finetuned and evaluated using a single NVIDIA Tesla V100 GPU. We restricted the selection further by only considering models that were either multilingual (including German) or trained exclusively on German language text. We expect that a Large Language Model, pretrained on German legal documents and finetuned for our task would achieve the best results possible.

**Bert-Based-German-Cased.** Based on the architecture of *BERT* [DEVLIN 2018], this open-source Transformer-based language model uses 110 million parameters and was trained on 2.3 billion tokens from German versions of Wikipedia, the EU Bookshop corpus, Open Subtitles, CommonCrawl, ParaCrawl and News Crawl [MDZ 2019]. The model was trained and evaluated using the Hugging Face transformers library [HUGGINGFACE 2024] (License: MIT). We used a batch size of 8 with a learning rate of 5e-5, 3 epochs and the AdamW optimizer ($\beta_1=0.9$, $\beta_2=0.999$). In the rare case that a single clause exceeded the maximum input length of 512 tokens, we truncated it.

**Ada.** This closed source model by OpenAI is based on *GPT-3*, which uses 175 billion parameters [FLORIDI 2020]. It was finetuned using the OpenAI API. The model has been deprecated on January 4, 2024 we still provide the results for comparison reasons.

**GPT-3.5.** The closed source model behind the first version of *ChatGPT* [OPENAI 2023]. We used the *gpt-3.5-turbo-1106* version. It is the only model we evaluated using both fine-tuning (via the OpenAI API) and prompt engineering.

**GPT-4.** The latest closed source model family by OpenAI [ACHIAM 2023]. We evaluated the *gpt-4-1106-preview* version as well as the by now newest version, *gpt-4o*.

## 5.2. Data Preprocessing

First, we removed all samples from the dataset which contained duplicate clauses. In case they differed, we decided to keep the more critical annotations, prioritizing void over unfair and unfair over valid. According to the annotating lawyers, overlooking problematic clauses is a bigger issue than an increased rate of false positives. The refined dataset contained 893 samples. Our experiments included different ways of constructing the model input, which included giving additional instructions or information to the model. The following transformations were used:

- **Clause Only:** *Ada* and *bert-based-german-cased* were initially trained and evaluated using only the clause itself as the input, meaning that the model neither had a meaningful textual instruction on what to do, nor the title of the clause's section as part of the input.

- **Clause & Title** [T]: We then tried to include the title of the clause into the input. While the title of a clause does not directly influence if the clause is void or not, the imbalance of the relative amount of problematic clauses across the different categories, visible in table 2 and closely related to the title, suggested that it could be an important indicator for the ML model. Under this assumption, we constructed the input as the concatenation of the title and the clause text (<title>: <clause>). The clause *'Über die Höhe des Einkommens oder sonstiger Bezüge ist - insbesondere auch den Mitarbeitern gegenüber - strengste Verschwiegenheit zu bewahren.'* with the title *'Vertragsstrafe/Wettbewerbsverbot'* would therefore be concatenated to *'Vertragsstrafe/Wettbewerbsverbot: Über die Höhe des Einkommens oder sonstiger Bezüge ist - insbesondere auch den Mitarbeitern gegenüber - strengste Verschwiegenheit zu bewahren.'*.

- **Clause & Instruction** [I]: In some experiments with OpenAI Large Language Models we provided instructions on what task to perform on the clause text. We did that by providing a 'system' message to the OpenAI API when fine-tuning and/or evaluating *GPT-3.5* and *GPT-4*, which reads as follows: *'Du bist ein deutscher Anwalt, welcher auf Wirtschaftsrecht spezialisiert ist und Klauseln aus Arbeitsverträgen auf ihre Zulässigkeit und Fairness bewertest. Du bewertest eine Klausel als proble-*

*matisch, wenn sie unfair oder nach deutschem Recht unzulässig ist. Andernfalls bewertest du sie als okay.'* (en: *'You are a German lawyer who specializes in commercial law and assesses the legality and fairness of clauses in employment contracts. You rate a clause as problematic if it is unfair or void under German law. Otherwise, you rate it as okay.'*). Since the clauses are written in German, we decided to provide German instructions. We also conducted experiments with English instructions and desired output labels but the differences were negligible.

- **Clause, Title & Instruction** [TI]: We also experimented with the combination of providing both section title and instruction.

Before applying any of these transformations, the dataset was randomly shuffled and split in a 9 to 1 ratio, using 90% of the data for training (whenever we finetuned models) and 10% for the evaluation. In the provided dataset the rows are already shuffled, with the first 90% of rows constituting the train set and the last 10% of rows building the test set. Our experiments included multiple runs using different prompts, splits etc. However, the reported results are all based on a single run on this split.

### 5.3. Results and Discussion

Table 3 shows the prediction performance measurements for our tested models. The arguably best results were achieved by fine-tuning the OpenAI *gpt-3.5-turbo-1106 [I]* model with the instruction given as a 'system' message and the pure clause (without the section title [T]) as the input. It achieved the highest F1-Score for the positive class (Void/Unfair) and therefore the best compromise between Precision and Recall, in which it was respectively outperformed by the finetuned open source model *bert-base-german-cased [T]* and the OpenAI prompt engineering model *gpt-3.5-turbo-1106 [TI]*. It also achieved the highest weighted average scores in all metrics.

For both fine-tuning and prompt engineering, it holds true that larger models generally produce better results. Surprisingly, providing the section title of the clause resulted in slightly worse performance in all fine-tuning experiments, while mostly yielding better results with prompt engineering. This even holds true for *gpt-3.5-turbo-1106* which was used in both scenarios. The drop in performance was most significant for the now deprecated *Ada* model where it led to a total collapse in precision for the Void/Unfair (1) class. Another interesting observation is the significant bias of OpenAI's models *gpt-3.5-turbo-1106*, *gpt-4-1106-preview* and *gpt-4o* towards the Void/Unfair (1) class, leading to high recall and low precision scores. The full-text answers of both models indicate that the models are inclined to position themselves as defenders of the employee, even though there were no such instructions in the provided prompt. As defenders of the employee, the models then suspect unfairness in many clauses that were classified as unproblematic by the annotating lawyers.

|  | Void/Unfair (1) | | | Weighted Average | | |
| --- | --- | --- | --- | --- | --- | --- |
| **Exact Model Version** | Precision | Recall | F1-score | Precision | Recall | F1-score |
| | *Finetuned Models* | | | | | |
| **ada** | 54.5% | 50.0% | 52.2% | 87.4% | 87.8% | 87.6% |
| **ada [T]** | 8.1% | 46.2% | 13.8% | 49.3% | 16.7% | 18.6% |
| **gpt-3.5-turbo-1106 [I]** | 61.5% | 61.5% | **61.5%** | **88.9%** | **88.9%** | **88.9%** |
| **gpt-3.5-turbo-1106 [TI]** | 57.1% | 61.5% | 59.3% | 88.2% | 87.8% | 88.0% |
| **bert-base-german-cased** | 66.7% | 30.8% | 42.1% | 86.0% | 87.8% | 85.8% |
| **bert-base-german-cased [T]** | **75.0%** | 23.1% | 35.3% | 86.4% | 87.8% | 84.9% |
| | *Prompt Engineering Models* | | | | | |
| **gpt-3.5-turbo-1106 [I]** | 15.9% | 84.6% | 26.8% | 79.7% | 33.3% | 37.0% |
| **gpt-3.5-turbo-1106 [TI]** | 17.3% | **100%** | 29.5% | 88.1% | 31.1% | 32.2% |
| **gpt-4-1106-preview [I]** | 25.8% | 61.5% | 36.4% | 82.0% | 68.9% | 73.2% |
| **gpt-4-1106-preview [TI]** | 28.6% | 76.9% | 41.7% | 85.0% | 68.9% | 73.5% |
| **gpt-4o [I]** | 27.9% | 92.3% | 42.9% | 87.8% | 64.4% | 69.7% |
| **gpt-4o [TI]** | 25.0% | 92.3% | 39.3% | 87.1% | 58.9% | 64.6% |

Table 3: Eval. of selected Closed- and Open-Source Models, Measured in Precision, Recall and F1-Score

## 6. Conclusion

### 6.1. Contribution

In this paper we present our efforts towards developing AI-assisted tools for reviewing German employment contracts using Natural Language Processing (NLP). Through a collaboration with a German law firm specializing in Economic Law, we have constructed a German dataset consisting of 1094 anonymized employment contract clauses, each annotated for legality and categorization. The inter-annotator agreement is 96.4% for legality labels and 100% for categories. This dataset serves as a valuable resource for future research in legal NLP, particularly for the German language. We provide baseline model benchmarks for the dataset, including both fine-tuning and prompt engineering approaches. Additionally, our experiments reveal a tendency for models, particularly those from OpenAI, to exhibit a bias towards protecting employees rather than employers, imposing a potential challenge for future applications.

### 6.2. Future Research

This paper, along with the released dataset and provided model benchmarks, marks the starting point for our contribution to AI-assisted German employment contract review. Over the course of this project, we aim to increase the dataset size to 10,000 clauses. Furthermore, we plan to design and evaluate advanced classification pipelines, leveraging RAG [GAO 2023], prompt engineering, and fine-tuning approaches. Ultimately, our efforts should lead to the design and evaluation (technical, economic, and social) of a prototypical AI-assisted employment contract review system. In doing so, we aim to bridge the gap between legal NLP research and legal practice, providing guidance for future AI usage in the legal domain.

## 7. Limitations

While the size of our dataset should be sufficient to evaluate pure prompt engineering approaches (according to the annotating lawyers, it covers the vast majority of clause types that appear in usual employment contracts), it might be a limiting factor for the performance of finetuned models. Since training and validation should not use the same samples, the dataset has to be split, potentially leading to an insufficient size of training and/or validation data. We will address this issue by increasing the dataset size to 10,000 clauses over the course of our collaborative project. Additionally, our model benchmarks serve only as baselines; they do not include extensive hyperparameter tuning and the number of tested prompts is also limited.

## 8. Ethical Considerations

**Privacy Protection:** All published data was thoroughly scanned for any personal or sensitive information including but not limited to names, contact details, addresses, date of birth, nationality and revealing job titles or team/department names. In most cases, personal details are only described at the start of an employment contract, and the two parties are thereafter referred to as the "employer" and the "employee," making further changes to any clauses unnecessary. In all other cases, personal information was pseudonymized by replacing it with imaginary names, addresses etc.

**AI-assisted Employment Contract Analysis:** As they are usually written by the employer, employment contracts with void clauses tend to disadvantage the employee. Due to insufficient legal knowledge and limited financial resources for professional legal advice, employees may unknowingly accept unfair working conditions. Conversely, if void clauses are identified, employers — who likely did not include these clauses intentionally — face the risk of costly lawsuits. Therefore, cheaper, faster, and more accessible legal analysis of employment contracts benefits both employees and employers, promoting greater transparency and fairness between the parties.